\documentclass[letterpaper, 10 pt, conference]{ieeeconf}
\IEEEoverridecommandlockouts

\usepackage{cite}
\usepackage{amsmath,amssymb,amsfonts}
\usepackage{algorithmic}
\usepackage{graphicx}
\usepackage{textcomp}
\usepackage{multirow}
\usepackage{xcolor}
\usepackage{leftidx}
\usepackage{color,soul}
\usepackage{caption}
\usepackage{subcaption}
\usepackage{makecell}
\usepackage[export]{adjustbox}
\usepackage{hyperref}
\usepackage{booktabs}
\usepackage[nolist]{acronym}
\usepackage{tikz}

\newacro{slam}[SLAM]{Simultaneous Localization and Mapping}
\newacro{uav}[UAV]{Unmanned Aerial Vehicle}
\acrodefplural{uav}[UAVs]{unmanned aerial vehicles}
\newacro{gns}[GNS]{Global Navigation Satellite}
\newacro{gnss}[GNSS]{Global Navigation Satellite System}
\acrodefplural{gnss}[GNSS]{Global Navigation Satellite Systems}
\newacro{mcl}[MCL]{Monte-Carlo localization}
\newacro{imu}[IMU]{Inertial Measurement Unit}
\newacro{dof}[DOF]{degree-of-freedom}
\acrodefplural{dof}[DOFs]{degrees-of-freedom}
\newacro{ransac}[RANSAC]{random sample consensus}
\newacro{map}[MAP]{maximum a posteriori}
\newacro{mle}[MLE]{maximum likelihood estimation}
\newacro{rms}[RMS]{root-mean-square}

\newcommand{\figref}[1]{\hyperref[#1]{Fig.~\ref*{#1}}}
\newcommand{\tabref}[1]{\hyperref[#1]{Tab.~\ref*{#1}}}
\newcommand{\secref}[1]{\hyperref[#1]{Sec.~\ref*{#1}}}
\newcommand{\algoref}[1]{\hyperref[#1]{Alg.~\ref*{#1}}}

\newcommand{\subfigref}[1]{(\subref{#1})}

\def\xycoords{$(x, y)$-coordinates}
\def\gnss{\ac{gnss}}
\def\gnssp{\acp{gnss}}
\def\ground{ground-truth}
\def\Ground{Ground-truth}
\def\bestcolor{(best viewed in color)}

\def\ie{\textit{i.e.},}

\newcommand\copyrighttext{%
  \footnotesize This work has been submitted to the IEEE for possible publication. Copyright may be transferred without notice, after which this version may no longer be accessible.}
\newcommand\copyrightnotice{%
\begin{tikzpicture}[remember picture,overlay]
\node[anchor=south,yshift=10pt] at (current page.south) {\fbox{\parbox{\dimexpr\textwidth-\fboxsep-\fboxrule\relax}{\copyrighttext}}};
\end{tikzpicture}%
}

\title{\LARGE \bf
GNSS-denied geolocalization of UAVs by visual matching of onboard camera images with orthophotos}

\author{Jouko Kinnari$^{1}$, Francesco Verdoja$^{2}$ and Ville Kyrki$^{2}$
\thanks{This work was supported by Saab Finland Oy.}
\thanks{$^{1}$J. Kinnari is with Saab Finland Oy,
Salomonkatu 17B, 00100 Helsinki, Finland
{\tt\small jouko.kinnari@saabgroup.com}}%
\thanks{$^{2}$F. Verdoja and V. Kyrki are with School of Electrical Engineering, Aalto University, Finland {\tt\small \{firstname.lastname\}@aalto.fi}}%
}

\begin{document}

\maketitle

\copyrightnotice

\begin{abstract}
Localization of low-cost \acp{uav} often relies on \acp{gnss}. \acp{gnss} are susceptible to both natural disruptions to radio signal and intentional jamming and spoofing by an adversary. A typical way to provide georeferenced localization without  \acp{gnss} for small UAVs is to have a downward-facing camera and match camera images to a map. The downward-facing camera adds cost, size, and weight to the UAV platform and the orientation limits its usability for other purposes. In this work, we propose a Monte-Carlo localization method for georeferenced localization of an \ac{uav} requiring no infrastructure using only inertial measurements, a camera facing an arbitrary direction, and an orthoimage map. We perform orthorectification of the \ac{uav} image, relying on a local planarity assumption of the environment, relaxing the requirement of downward-pointing camera. We propose a measure of goodness for the matching score of an orthorectified \ac{uav} image and a map. We demonstrate that the system is able to localize globally an \ac{uav} with modest requirements for initialization and map resolution.
\end{abstract}

\acresetall

\section{Introduction}

Geolocalization---finding the Earth-fixed coordinates---of an \ac{uav} in outdoor scenarios typically relies on \gnssp{}. \gnssp{} are naturally susceptible to blockages and reflections in radio signal path and intentional jamming or spoofing by an adversary. For these reasons, especially in critical security and logistics applications, one cannot rely only on \gnss{} as the sole source of geolocalization of an autonomous vehicle.

A high interest in the recent decades has existed on \ac{slam} solutions~\cite{7747236}. In \ac{slam}, a sensor system is used to build a map on the go and localize the vehicle with respect to the self-built map. A \ac{slam} system cannot, however, provide Earth-fixed coordinates without georeferenced landmarks or other localization infrastructure, and odometry error accumulation can only be partly compensated if the vehicle traverses an area it has visited previously.

An alternative to \ac{slam} is to match observations of the vehicle to a separately acquired map. While this approach requires that a map of the environment is available, the benefits include the ability to provide Earth-fixed coordinates, tolerate uncertainty in starting pose and compensate for odometry drift over long travel distances. Another challenge of localization using a pre-acquired map is that the sensor system with which the map is collected is typically different from the sensor system onboard an \ac{uav}, and matching the sensor observations to the map is not a trivial task.

\begin{figure}
  \includegraphics{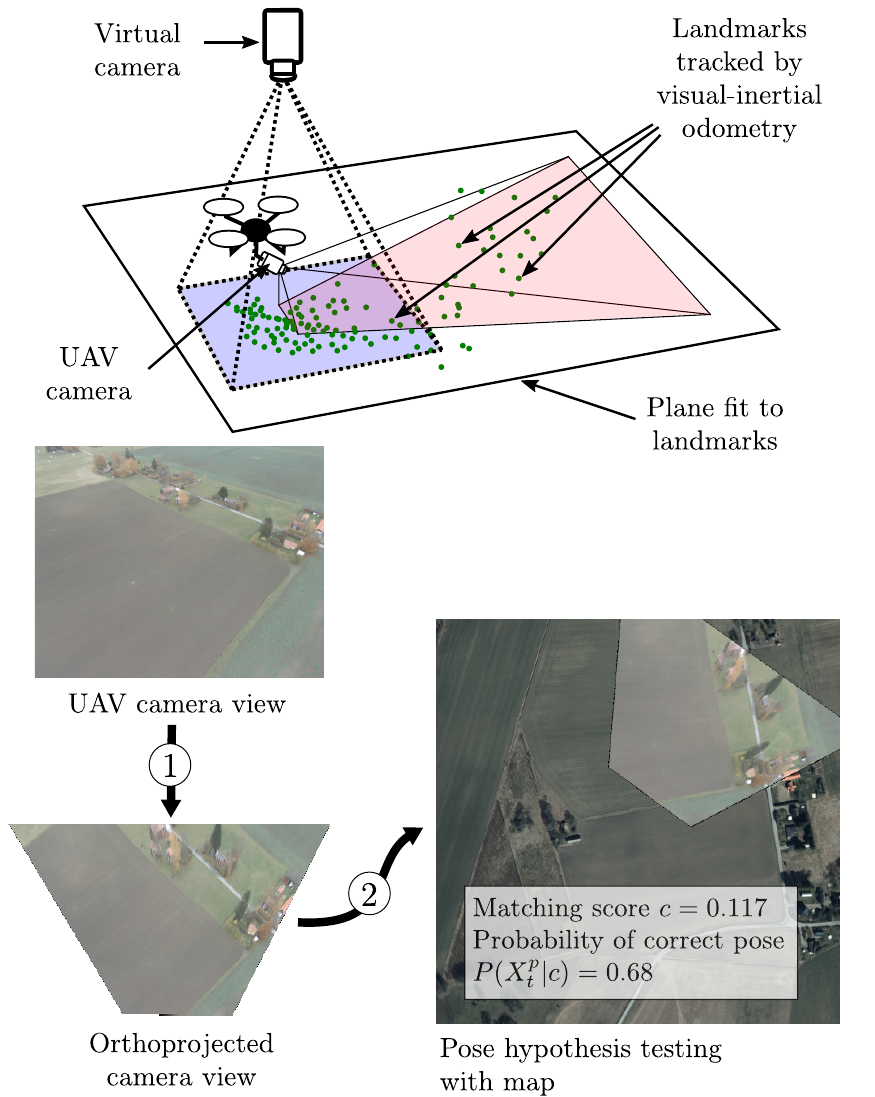}
  \caption{Proposed geolocalization framework: Visual-inertial odometry is used for tracking \ac{uav} motion and visual landmarks, and solving landmark 3D positions with respect to \ac{uav} pose. \ac{uav} camera image is projected through planar homography to a virtual view looking straight down. The orthoprojected image is used for map matching and pose tracking in a particle filter framework.} \label{fig:overview}
\vspace{-1em}
\end{figure}

In this paper, we address these problems and present a method able to perform absolute visual localization by tracking \ac{uav} motion via visual-inertial odometry, orthorectifying \ac{uav} images, matching the orthorectified images to a georeferenced orthoimage map, and fusing measurements with a particle filter. An illustration of the main parts of the system is shown in \figref{fig:overview}.

Novel contributions in this work are threefold. Firstly, we propose a localization solution which does not put strict requirements on \ac{uav} camera orientation. This relaxes the requirement of having a downward-facing camera, which is a common requirement in \ac{uav} visual localization systems. This enables the use of a single camera in an \ac{uav} for not only localization, but also other tasks related to the primary mission of the autonomous agent with no requirement for intermittently orienting camera downwards.

Secondly, we compare a number of classical image matching metrics in \ac{uav} localization setting and pinpoint the characteristics of a good matching quality metric.

Thirdly, we demonstrate the convergence of the position estimate of a localization system in a setting where only very limited amount of information is available on true starting pose. We demonstrate the end-to-end operation of the proposed system with three datasets acquired for this task, using a map with a modest 1 m per pixel resolution, and compare localization results to visual-inertial odometry.

\section{Related work}

A number of recent works have presented ways to implement \ac{uav} geolocalization without satellite navigation~\cite{COUTURIER2021103666}. A common solution for providing ground matches consists of a camera pointed approximately downward. By having a downward-looking camera at a high flight altitude, the camera images appear similar to orthophotos or satellite images. In this way, the problem of \ac{uav} localization approaches image template matching. A large number of publications~\cite{9341682, 10.1117/12.2518516, MANTELLI2019304, schleiss2019translating, 7418753, 6942633, 6521919, 6943040} use a downward-pointing camera. Requiring such a setup is a major limitation: either a downward-pointing camera is added for the sole purpose of localization of an \ac{uav}---which adds cost, size, weight, and power requirements---or the mission camera must, at least at intervals, be oriented downwards, thereby intermittently limiting the availability of that camera for the primary task of the \ac{uav}.

A number of \ac{uav} image-to-map matching approaches use image features~\cite{6942633} or semantic features~\cite{5354165}. Our work focuses on area-based matching solutions with the motivation that using the full image area of an \ac{uav} image instead of sparse feature points provides opportunity to utilize all available information in an image for this task.

Other localization solutions relying on using an estimated 3D structure and matching it to a previously acquired 3D structure of the same environment exist~\cite{6521919}, as well as using a 3D model of the environment for rendering images of a planned trajectory, and then finding the pose of an \ac{uav} based on best-matching rendered image~\cite{9196606}. However, acquiring a suitably fine-grained up-to-date 3D map can be laborious, costly or even impossible, compared to having orthophotos or satellite images of the area in which the \ac{uav} is flown.

The requirements on the amount of information needed for initialization vary in different works. A common assumption is that \ac{uav} starting pose has to be known accurately~\cite{9196606, 8794228, 8575361} for the system to work. In other works~\cite{MANTELLI2019304, 9341682} no information on the initial pose is required, but the size of the map in which the \ac{uav} is expected to operate is relatively small (1.1 to 5.7 km$^2$). To make a balanced choice for initialization requirements and map size, in this work we refer to an expected use case for an autonomous \ac{uav}, assuming that a user is able to state a 200$\times$200 m area in which the flight mission starts, with no information on orientation, and assuming no limitations on the maximum map size.

A key challenge in \ac{uav} localization is appearance change due to weather, illumination, and season. The choice of the matching method of an observation of an \ac{uav} to a base map is not trivial. The matching criteria used range from classical computer vision image registration methods including mutual information and derivatives of it~\cite{6943040, 9196606} to custom measures tailored for this problem, often using various deep learning-based methods~\cite{9341682, 8575361, schleiss2019translating, 8793558, 8794228}. We experimentally compare a number of classical area-based matching criteria to find the best matching means suitable for this purpose, and demonstrate a systematic way to assess the suitability of a matching criterion for this task.

\section{Methodology}\label{sec:methods}

\begin{figure*}
\centering
  \includegraphics[width=0.9\textwidth]{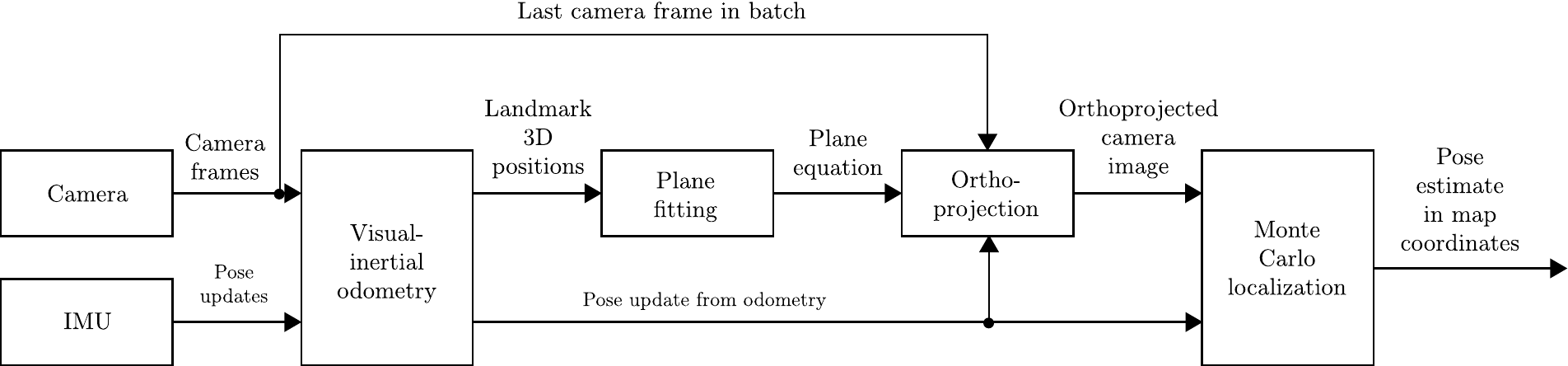}
  \caption{Block diagram of proposed solution}
  \label{fig:block_diagram}
  \vspace{-1em}
\end{figure*}

The method we present for localizing an \ac{uav} by matching an orthorectified \ac{uav} image to a map consists of a number of subcomponents. An illustration of the subcomponents of the proposed solution are shown in \figref{fig:block_diagram}.

First, the \ac{uav} camera frames and \ac{imu} measurements are fused in a visual-inertial odometry system. Observations of 3D coordinates of landmarks are used in estimating a locally planar model of the ground under the \ac{uav}. With the model of the ground, orthoprojection of the camera image is done. That orthoprojection, along with measurements of egomotion of the \ac{uav} based on visual-inertial odometry, are then fused in a \ac{mcl} framework~\cite{ThrunProbabilisticRobotics05}. For using the orthoprojected \ac{uav} image in \ac{mcl}, we use a method of matching the image to a map.

We describe the implementation of each subcomponent, starting from \ac{mcl}, continuing to visual-inertial odometry, and finally we look at generating the orthoprojection and matching the orthoprojection to a map.

\subsection{\acl{mcl}} \label{sec:mcl}

\ac{mcl} is an adaptation of particle filtering to the localization problem. We first describe the state of the system which we want to estimate by filtering, then describe means for initializing the filter and finally describe some characteristics of how prediction and particle weighing are performed in our work. The choice of using particle filters for global fusion has precedent in other previous works on vision-based localization~\cite{10.1117/12.2518516, MANTELLI2019304, 7418753}.

\subsubsection{Definition of state}

The full search space of 3-D localization is 6-dimensional, consisting of the 3-\ac{dof} position and 3-\ac{dof} orientation of the \ac{uav} with respect to an Earth-fixed coordinate frame. By using an \ac{imu}, we can simplify the problem by two degrees of freedom (roll and pitch) since the direction of gravity can be estimated.

We will also not directly solve for the altitude of the \ac{uav}. The map which we are using for global localization contains only 2D information. Based on the map we are thus not able to directly infer the altitude of the \ac{uav}. However, with the proposed method we are able to compute the vertical distance from the \ac{uav} to a locally planar model of the ground.

In visual-inertial odometry, scale exhibits drift.
In order to tolerate the drift, we define a scale parameter $s$ which we also infer as part of the search algorithm.

Thereby we formulate the \ac{mcl} search problem as the problem of estimating state $X_t$ at time $t$:
\begin{equation}
    X_t = (x(t),y(t),\phi(t),s(t))\enspace,
\end{equation}
where $x(t)$, $y(t)$ are longitude and latitude of the \ac{uav} position in map coordinate system, $\phi(t)$ is the yaw (heading) of the \ac{uav} and $s(t)$ is the scale parameter at time t. 

Following the particle filter algorithm and low variance sampler described in~\cite{ThrunProbabilisticRobotics05}, we represent our belief of estimated state $\hat{X}_t$ with a set of $P$ particles as $\hat{X}_t = \{X^{p}_t\}$, $p=0 \ldots P$, where
\begin{equation}
    X^{p}_t = (x^{p},y^{p},\phi^{p},s^{p},w^{p})\enspace,
\end{equation}
$X^p_t$ thus represents one particle with index $p$ at time $t$. The weight of each particle is denoted $w^{p}$. We resample after each step. In experiments, we use $P=1000$.

\subsubsection{Initialization}

We assume that the user of the \ac{uav} is able to infer the starting latitude-longitude coordinates on a map at a limited accuracy, and that inferring the initial heading of an \ac{uav} is a more tedious and error-prone task. With this motivation, we initialize the particle filter's \xycoords{} such that the prior distribution is assumed to be uniform over a rectangular area of $d \times d$ meters, with no prior information on heading. In experiments, we set $d=200$ m.

\subsubsection{Prediction and particle weighing}

Prediction of particle movement is based on visual-inertial odometry, described in detail in \secref{sec:vio}. Updating particle weights is based on matching an orthorectified image to a map, detailed in \secref{sec:mapmatching}.

\subsection{Visual-inertial odometry} \label{sec:vio}

Visual-inertial odometry consists of three subparts: detection and tracking of features (landmarks) on the image plane in each camera image, measurements from the \ac{imu} of the \ac{uav}, and fusion of these separate measurements.

\subsubsection{Feature detection and tracking}

The visual odometry front-end consists of detecting good features to track using the Shi-Tomasi detector~\cite{shi1994good} and tracking their movement across consecutive image frames using a pyramidal Lucas-Kanade tracker~\cite{bouguet2001pyramidal}, both available in OpenCV~\cite{opencv}. The pixel coordinates $z_{i,m}=[u_{i,m},v_{i,m}]$ of each feature, indexed by $i$, in each frame, indexed by $m$, are recorded. Each tracked feature corresponds to a landmark in the scene, and each landmark is assumed to be static, \ie{} not moving with respect to other landmarks. To improve tracking of features in situations of significant rotation of the camera, we initialize the optical flow tracker with a rotation-compensated initial guess for feature pixel coordinates computed from feature coordinates of previous image, where the rotation estimate comes from the \ac{imu} measurement.

To filter out faulty correspondences of landmark detection between frames, a \ac{ransac}-based fundamental matrix estimation step is taken to exclude outliers from following steps. The estimated fundamental matrix is not utilized beyond outlier removal.

\subsubsection{Inertial odometry}

We assume that an inertial odometry solution, capable of producing translation and rotation of camera poses between frames and the associated uncertainty of the inter-pose transformation from the measurements of an \ac{imu} device, is available. We also assume that the \ac{imu} is able to infer direction of gravity at a high accuracy. In this paper we do not illustrate in detail the implementation of such system and an interested reader is referred to~\cite{titterton2004strapdown}.

\subsubsection{Coordinate frame conventions}

To formulate the visual-inertial odometry solution, we define a number of coordinate systems and transformations.

A coordinate system $\{C_m\}$ is tied to the $m$-th camera pose such that camera center is at origin, principal axis is in positive $z$-axis direction, $x$-axis points right along camera image plane, and $y$-axis is perpendicular to both, forming a right-handed coordinate system.

We want to define a transformation from the camera coordinate system $\{C_m\}$ to another coordinate system, where one of the coordinate axes, $z$, is aligned with the direction of gravity. In this way, $z$ component in the new frame is altitude. In our new coordinate frame, we select $x$ coordinate axis to point in the direction of heading of the camera in a given frame $m$. We therefore define a coordinate system $\{B_m\}$ where $z$-axis points opposite to direction of gravity $g$. Positive $x$-axis in frame $\{B_m\}$ is chosen to point in direction of camera principal axis in frame $\{C_m\}$, while being perpendicular to $z$-axis. $y$-axis is perpendicular to both $z$- and $x$-axes, forming a right-handed coordinate system.

In this way, the coordinate frame computed for the first frame in a batch, $\{B_0\}$, is such that negative $z$ of $\{B_0\}$ corresponds to gravity and $x$ of $\{B_0\}$ is towards camera optical axis of the first frame in batch but perpendicular to gravity, and the origin is at the camera center of the first frame.

\subsubsection{IMU noise model}\label{sec:imunoisemodel}

We define inertial measurements as inter-pose transformations corrupted by noise $n_m$. We assume $n_m$ to be Gaussian zero-mean noise, defined in the Lie algebra of $\mathbb{SE}(3)$ with a diagonal covariance matrix in which each position coordinate component has variance $\sigma_p^2(\tau)$ and rotation component has variance $\sigma_\omega^2(\tau)$, where $\sigma_\omega(\tau)$ is the standard deviation of angular random walk over a period of $\tau$, which we compute as~\cite{woodman2007introduction}
\begin{equation}
    \sigma_\omega(\tau) = N_\omega \cdot \sqrt{\tau}\enspace,
\end{equation}
where $N_\omega$ is a characteristic of the gyroscope used in the system. Similarly, the standard deviation of random walk on position error components is computed as~\cite{woodman2007introduction}
\begin{equation}
    \sigma_p(\tau) = N_v \cdot \tau^{\frac{3}{2}}\enspace,
\end{equation}
where $N_v$ is a characteristic of the accelerometer used in the system. In demonstrations, we use $N_v = 0.029  m/s/\sqrt{h}$ and $N_\omega = 0.26 ^o/\sqrt{h}$ based on a reference component~\cite{ADIS16488Adatasheet}.

\subsubsection{Fusion of \ac{imu} poses and feature locations}

Fusing inertial measurements with static feature detections is based on \ac{map} estimation of a batch of camera poses.

To define the number of camera frames to include in a batch, we use simple heuristic rules: we require that a minimum of 100 \emph{valid} features have been observed, and that the \ac{uav} has traveled a minimum of 100 m since last batch computation (based on inertial odometry); a feature is considered valid if, during the time it was tracked, the \ac{uav} traveled a minimum of 20 meters, based on inertial odometry.

For the full batch of images and feature (landmark) coordinate observations, we solve for landmark 3D coordinates and camera a posteriori pose and pose covariance for each camera frame in batch similarly to~\cite{doi:10.1177/0278364906072768} with the exception that we add a Huber loss~\cite{huber1964} to the projection error using parameter value $\delta=10.0$ to gain additional robustness to feature correspondence outliers that the earlier \ac{ransac} outlier rejection step did not exclude. The \ac{map} optimization problem is solved numerically using GTSAM software package~\cite{gtsam}, defining the cost function as a factor graph.

The optimization is initialized with mean pose estimated by the \ac{imu} and computing an initial estimate for landmark coordinates using the linear triangulation method in~\cite{HARTLEY1997146}. A Dogleg optimizer with GTSAM's default parameters is used.

\subsubsection{Prediction in \ac{mcl}}

We use the \ac{map} estimate and marginal covariance matrix of the pose of the last camera frame in a batch for drawing samples for odometry prediction in \ac{mcl}. The variance of scale between batches is assumed to be 0.01.

\subsection{Generating orthoprojection from camera image}

One result of fusion of \ac{imu} poses and feature locations is an estimate of landmark position mean values $l_i$ for each landmark. We assume that each detected landmark belongs to static, locally flat terrain, and find a plane $q$ that best fits to the landmarks observed in the batch by least squares fitting. We then use the equation of that plane for projecting an orthographic projection of the camera image of the last frame in the batch.

For each pixel coordinate $v_c \in \{v_{ul}, v_{ur}, v_{ll}, v_{lr}\}$, corresponding with upper left $v_{ul}$, upper right $v_{ur}$, lower left $v_{ll}$, lower right $v_{lr}$ image corner pixel coordinate, we project a ray from the camera center through that pixel and solve the point in which it intersects plane $q$. We denote the intersection points $p_{ul}$, $p_{ur}$, $p_{ll}$ and $p_{lr}$, respectively, stated in frame $\{B_m\}$.

\begin{figure}
\centering
\includegraphics{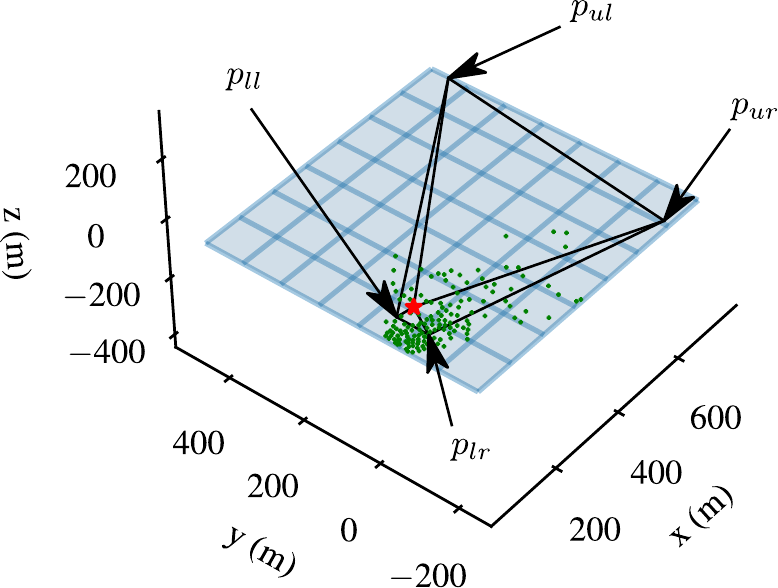}
\caption{Feature point locations (green dots), plane fit to points (blue grid), camera center (red star), and intersection points of rays from camera center through image corner coordinates with plane ($p_{ll}$, $p_{ul}$, $p_{ur}$, $p_{lr}$) for last camera pose in batch, shown in frame $\{B_0\}$ \bestcolor{}.}
 \label{fig:3d_image}
\end{figure}

\begin{figure}
    \centering
     \begin{subfigure}[b]{0.55\linewidth}
         \centering
         \includegraphics[height=3.7cm]{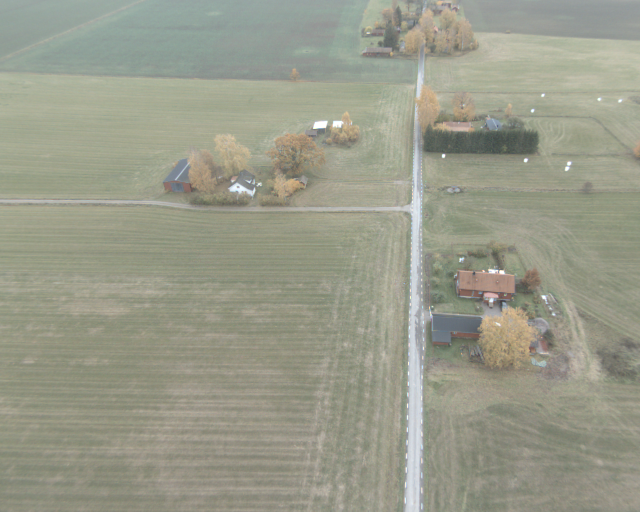}
         \caption{}
         \label{fig:undistorted_original_image}
     \end{subfigure}%
    \hfill 
    \begin{subfigure}[b]{0.42\linewidth}
         \centering
         \includegraphics[height=3.7cm]{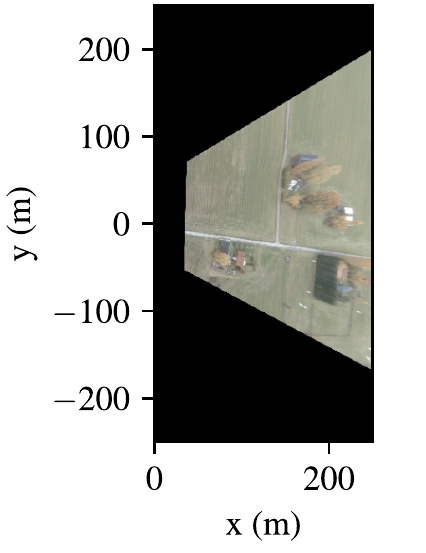}
         \caption{}
         \label{fig:orthoprojection_example}
     \end{subfigure} 
    \caption{Original camera image \subfigref{fig:undistorted_original_image} and orthoprojection \subfigref{fig:orthoprojection_example} of the same image.}
    \label{fig:uav_image_and_orthoprojection}
    \vspace{-1em}
\end{figure}

We then take the \xycoords{} of each corner intersection point from vectors $p_c \in \{p_{ul}, p_{ur}, p_{ll}, p_{lr}\}$ and find a homography matrix that transforms the pixel coordinates in camera image to the \xycoords{} in frame $\{B_m\}$ and use that homography matrix for transforming the last camera image to an orthoprojection $\Omega$ using bilinear interpolation. The orthoprojection is stored at a resolution of 1.0 m per pixel and $\Omega$ spans a range $[-250, 250]$ meters in $x$ direction and $[-250, 250]$ meters in $y$ direction. Besides the orthoprojection, also a mask image $\Omega_m$ defining valid pixels on $\Omega$ is generated. An illustration of feature points and intersection points $p_c$ is shown in \figref{fig:3d_image}, while an example of the original \ac{uav} image and the image after orthoprojection is presented in \figref{fig:uav_image_and_orthoprojection}.

\subsection{Image matching}\label{sec:mapmatching}

\subsubsection{Matching score}

The camera image obtained through orthoprojection provides measurement data that can be used for checking correctness of a state hypothesis. We measure the correctness of a state hypothesis $(x, y, \theta, s)$ by computing a matching score of image $\Omega$ to a map $\mathcal{M}$ using a suitable matching function. We denote such matching function as $c(x,y,\theta,s,\Omega,\Omega_m,\mathcal{M})$. The matching function scales and rotates image $\Omega$ and mask $\Omega_m$ according to scale parameter $s$ and heading $\theta$, takes a subimage from map $\mathcal{M}$ at 1 m/pixel resolution that corresponds with translation hypothesis $(x, y)$ and computes a (masked) matching score between the image $\Omega$ and subimage of map.

To find a suitable measure of correctness of match, in \secref{sec:exp}, we evaluate a number of classical correlation-based image similarity measures.

\subsubsection{From matching score to probability of correct pose}

We want to know the probability that evidence $c$ observed by the \ac{uav} is consistent with a state hypothesis $X^{p}_t$ and use that in particle weighting. We denote this probability as $P(X^{p}_t \mid c)$.

For the values of $c$ corresponding with \ground{} camera poses, we find a nonparametric probability density function $p(c \mid X^{p}_t)$ by computing a Gaussian kernel density estimate, using Scott's rule~\cite{doi:https://doi.org/10.1002/9780470316849} as bandwidth estimation method. Similarly, we find the probability density function $p(c \mid \neg X^{p}_t)$ for randomly drawn camera poses.

There is a chance, which we quantify with $\omega=0.1$, that we observe a matching score value that is not consistent with either distribution $p(c \mid X^{p}_t)$ or the distribution $p(c \mid \neg X^{p}_t)$. The reason for this may be for instance that the \ac{uav} is flown in an area that our data did not represent well. In order to allow existence of outliers, we also determine an outlier distribution $p(c \mid o)$. We assume that the outlier distribution is uniform across the range of values of $c$ observed in the data.

\begin{figure*}
 \centering
 \begin{subfigure}[t]{.32\linewidth}
 \centering
 \includegraphics[width=\linewidth]{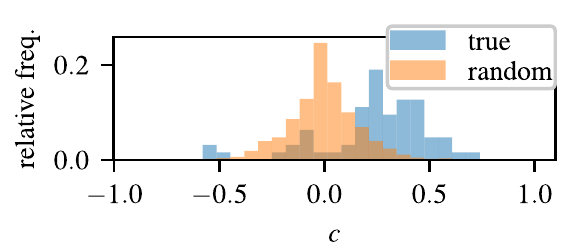}
 \caption{Values of $c$ computed for correct and randomly selected poses}
 \label{fig:histogram_matching_score}
 \end{subfigure}\hfill
 \begin{subfigure}[t]{.32\linewidth}
 \centering
 \includegraphics[width=\linewidth]{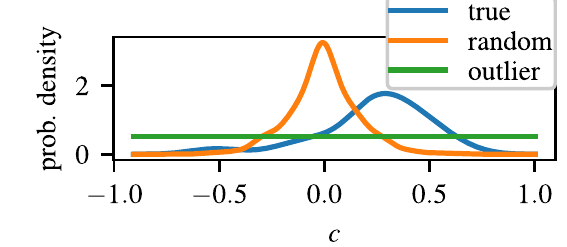}
 \caption{Probability density for different classes}
 \label{fig:probability_density_matching_score}
 \end{subfigure}\hfill     
 \begin{subfigure}[t]{.32\linewidth}
 \centering
 \includegraphics[width=\linewidth]{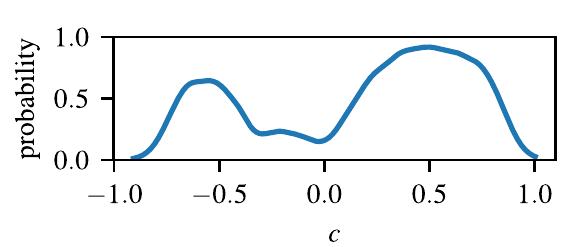}
 \caption{Probability of true pose, given matching score $c$.}
 \label{fig:probability_matching_score}
 \end{subfigure}
 \caption{Characterization of one of the matching scores: Moravec}
 \label{fig:matching_score_probability_plots}
\end{figure*}

Once we have the distributions for $p(c \mid X^{p}_t)$, $p(c \mid \neg X^{p}_t)$, and $p(c \mid o)$, we can compute the probability at which a matching score was drawn from the true match class:
\begin{equation}
    P(X^{p}_t \mid c) = \frac{p(c \mid X^{p}_t)}{p(c \mid X^{p}_t) + p(c \mid \neg X^{p}_t) + \omega p(c \mid o)}\enspace.
\end{equation}
We use the function $P(X^{p}_t \mid c)$ as the weighing function for each particle $p$ in the \ac{mcl} filter. In all experiments in \secref{sec:exp}, we identify $P(X^{p}_t \mid c)$ using dataset 1 and use that as weighing function. The histogram of values for $c$, the associated probability density functions and weighing function are illustrated in \figref{fig:matching_score_probability_plots}.

\section{Experiments}\label{sec:exp}

In order to evaluate the proposed localization solution, we study its performance on three datasets collected for the task. Firstly, we experiment with different matching criteria after orthorectification to find the best means for matching an orthorectified UAV image to a map.
Secondly, we assess the end-to-end localization performance of the proposed solution, when the system is initially given a 200 m by 200 m area in which the flight is expected to start, without information on heading.
Thirdly, we assess the end-to-end localization performance with perfect initialization, and compare localization results to a solution utilizing only odometry.

\subsection{Datasets}

\subsubsection{\ac{uav} data}

\begin{table}
\caption{\label{tab:dataset_characteristics}Characteristics of the datasets. Trajectory lengths were computed along $(x, y)$ plane. Camera angles are measured between nadir and camera principal axis.}
\begin{tabular}{p{3mm} p{6mm} p{8mm} p{6mm} p{24mm} p{13mm}} 
\toprule
Set & Area & Traj. length (m) & Alt. (m) & Mean camera angle [range] ($^{\circ}$) & Acquisition time\\
\midrule
1 & A & 6888 & 92 & 50.9 [48.5, 61.6] & Oct 2019 \\ 
2 & B & 4080 & 91 & 60.7 [55.2, 70.0] & Nov 2019 \\  
3 & B & 6361 & 92 & 52.6 [48.9, 118.9] & Nov 2019 \\ 
\bottomrule
\end{tabular}
\end{table}

We demonstrate the operation of our localization method using three datasets\footnote{Data provided by Saab Dynamics Ab.} that were collected using a commercial off-the-shelf hexacopter \ac{uav} with a custom data collection payload. The datasets consist of RGB images and their \ground{} position and orientation information, recorded at 10 Hz. Some characteristics of the datasets used in this study are given in \tabref{tab:dataset_characteristics}. In all flights, the drone started from ground, ascended to a set altitude above starting location, and was then flown at relatively constant altitude for the full trajectory. The camera frames are undistorted to correspond with a pinhole camera model with calibrated intrinsic parameters, and scaled to resolution $640 \times 512$ pixels.

\begin{figure}
    \centering
     \begin{subfigure}[t]{0.49\linewidth}
        \centering
         \includegraphics[width=\linewidth]{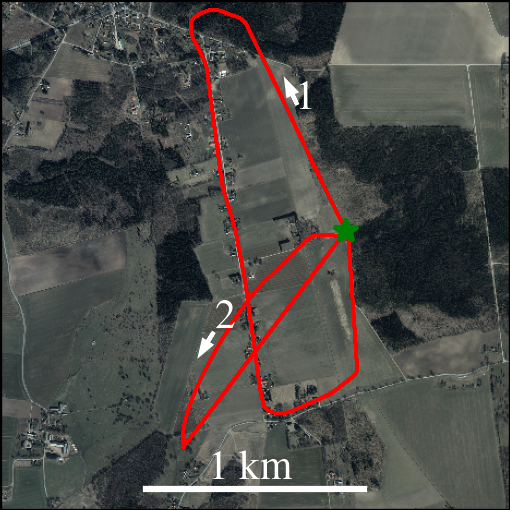}
         \caption{Map of area A (Klockrike, Sweden) and trajectory of dataset 1.}
         \label{fig:area_a_map}
     \end{subfigure}%
     \hfill
     \begin{subfigure}[t]{0.49\linewidth}
        \centering
         \includegraphics[width=\linewidth]{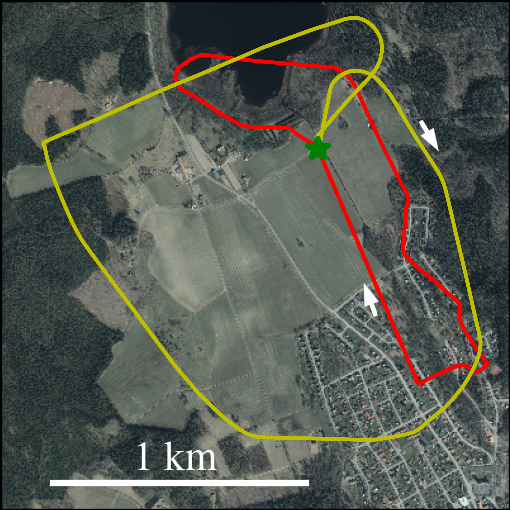}
         \caption{Map of area B (Kisa, Sweden) and trajectory of datasets 2 (red), 3 (yellow)}
         \label{fig:area_b_map}
     \end{subfigure}
     \caption{\Ground{} data. Starting location marked with green star. In both areas, the \ac{uav} flies over forest areas, fields, and residential areas as well as a lake in datasets 2 and 3 \bestcolor{}.}
     \label{fig:maps_and_trajectories}
\end{figure}

The \ground{} position and orientation of the camera for each frame were originally smoothed from RTK-GPS and \ac{imu} measurements using a proprietary smoothing algorithm. The \ground{} trajectories of all datasets are shown in \figref{fig:maps_and_trajectories}.

\subsubsection{Simulating noisy inertial measurements}

At the time of running the experiments, the originally acquired IMU data was not available. For that reason, we use \ground{} pose information to generate simulated \ac{imu} pose increments. We compute the \ground{} rotation and translation increments and simulate the effect of measurement noise by adding random noise according to \secref{sec:imunoisemodel}. In this way, we are able to simulate the impact of IMU measurement inaccuracies without undermining the purpose of the experiment.

\subsubsection{Maps}

As maps, we use georeferenced orthophotos, dated April 2019, which we purchased from a local map information supplier\footnote{Lantmäteriet, \url{https://www.lantmateriet.se/}.}, and scaled down to a resolution of 1 m/pixel.

\subsection{Matching criteria selection and weighting function characterization}

To measure the correctness of a pose hypothesis, given an orthorectified \ac{uav} image and a map, we compare a number of cross correlation-based matching methods. We implement all methods listed in~\cite{Roma2002}. We want to identify the distribution of matching scores for both correct hypotheses (using \ground{} pose data) and randomly drawn pose hypotheses, following the process described in \secref{sec:mapmatching}.

Optimally, the true and randomly-drawn distributions would not overlap. In such a case, the filter could immediately discriminate between correct and incorrect hypotheses. We use the amount of overlap as criterion for selecting a suitable matching score. To get a measure of overlap of the two distributions, we compute an overlapping coefficient, $o_c$, from the histograms of values of $c$ as
\begin{equation}
    o_c = \sum_{n=0}^{N} \min(h_t[n],h_g[n]) \cdot b\enspace,
    \label{eq:measure_of_overlap_of_two_distributions}
\end{equation}
where $h_t[n]$ and $h_g[n]$ are the relative frequencies of each value of $c$ in bin $n$, respectively, $b$ is the bin width in the histograms and $N$ is the number of bins in the histograms. The same binning is used for both histograms. We use $N=30$, and bin widths are selected so that they cover the full range of values observed of $c$. With this measure, the smaller the value of $o_c$, the less overlap there is between the true and generated distributions and the better the matching method is assumed to separate between true and generated matches.

\begin{table}
\centering
\caption{\label{tab:matching_scores}Overlapping coefficient computed for each matching score. Lower is better, the best performing method is indicated in bold. Method descriptions in \cite{Roma2002}.}
\begin{tabular}{l r} 
\toprule
Method & $o_c$ \\
\midrule
Sum of absolute differences & 0.606 \\
\midrule
Sum of squared differences & 0.556 \\
\midrule
Simple cross-correlation & 0.856 \\
\midrule
Normalized cross-correlation & 0.803 \\
\midrule
Zero-normalized cross-correlation coefficient & 0.357 \\
\midrule
Normalized Zero Mean Sum of Squared Differences & 0.905 \\
\midrule
Moravec & \textbf{0.345} \\
\midrule
Normalized sum of squared differences & 0.583 \\
\midrule
Zero Mean Sum of Squared Differences & 0.675 \\
\midrule
Zero mean sum of absolute differences & 0.646 \\
\midrule
Locally scaled sum of squared differences & 0.656 \\
\midrule
Locally scaled sum of absolute differences & 0.617 \\
\bottomrule
\end{tabular}
\end{table}

\begin{figure*}
 \centering
 \begin{subfigure}[b]{.32\linewidth}
 \centering
 \includegraphics[width=\linewidth]{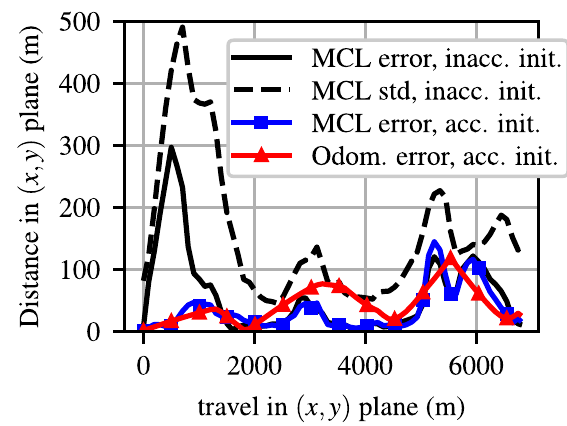}
 \caption{Dataset 1}
  \end{subfigure}\hfill
 \begin{subfigure}[b]{.32\linewidth}
 \centering
 \includegraphics[width=\linewidth]{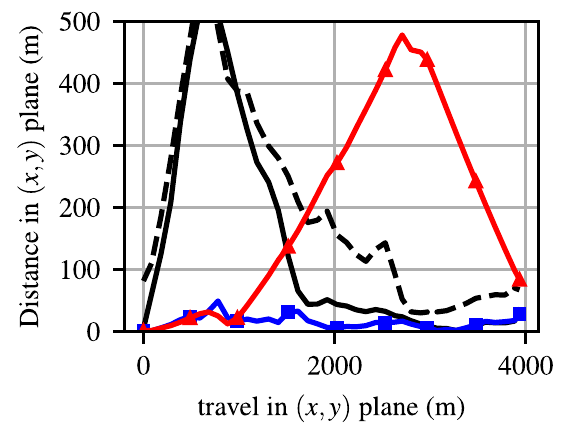}
 \caption{Dataset 2}
  \end{subfigure}\hfill     
 \begin{subfigure}[b]{.32\linewidth}
 \centering
 \includegraphics[width=\linewidth]{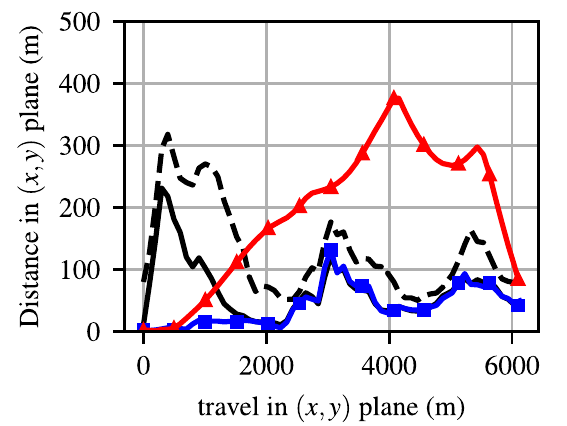}
 \caption{Dataset 3}
  \end{subfigure}
 \caption{Results from localization experiments}
 \label{fig:error_plots}
 \end{figure*}
 
We compute the overlapping coefficients $o_c$ using equation \eqref{eq:measure_of_overlap_of_two_distributions} for a number of matching methods using dataset 3 and assemble the results in \tabref{tab:matching_scores}. Based on the values of overlapping coefficient using different methods, we select Moravec as our matching method in localization performance experiments. For that matching method, we quantify the weighing function. The distribution of matching scores and weighing function is illustrated in \figref{fig:matching_score_probability_plots}.

\subsection{Localization performance}

Finally, we want to evaluate the ability of the proposed solution to perform global localization. To this end, we compute the weighted mean of the \xycoords{} in particle set $\hat{X}_t$. We compute the difference of weighted mean to \ground{} mean \xycoords{} and use it for assessing localization performance. To provide a measure of spread of $\hat{X}_t$, we also compute the weighted standard deviation of particle \xycoords{}. In order to evaluate the effect of the initialization procedure on the performance, we evaluated our method both with perfect initialization and the imprecise initialization described in \secref{sec:mcl}. To provide a point of comparison, we also plot the mean error of solution relying only on visual-inertial odometry.

\Ac{rms} error for each case is presented in \tabref{tab:rms_errors}, while in \figref{fig:error_plots} we show the profiles of the localization error over the length of the flight. In the plot it is evident that, if the starting pose is known perfectly when the \ac{uav} starts the mission, the \ac{mcl} solution is able to produce a smaller mean error than an odometry solution. When considering inaccurate initialization instead, we see that with each dataset, the system initially diverges significantly. This is due to unknown starting heading: the particles representing incorrect headings survive for a period of time after starting the filter; however, after a couple of kilometers of travel, those particles die out because of poor match of \ac{uav} image to map. After convergence, localization error is on a par with the solution that was given perfect information on initial starting pose. This shows that the proposed system is able to provide global localization even under significant uncertainty in the initial pose.

It is worth noting that by quantifying positioning error as Euclidean distance between estimated and true pose, errors in heading estimation show as an increased error in $(x, y)$ plane when the distance of the \ac{uav} increases from the starting position. This explains the apparent difference in odometry-based positioning error across different datasets.

\begin{table}
\centering
\caption{\label{tab:rms_errors}\ac{rms} errors in meters on different datasets, with different initializations, and comparison to odometry error. Lower is better.}
\begin{tabular}{p{0.7cm} p{0.1\textwidth} p{0.1\textwidth} p{0.1\textwidth}} 
\toprule
Dataset & \ac{mcl} & \ac{mcl} & Odometry \\
& inaccurate init & accurate init & accurate init \\
\midrule
1 & 87.54 & 48.21 & 51.53 \\
2 & 213.23 & 17.85 & 252.84 \\
3 & 80.57 & 51.07 & 216.98 \\
\bottomrule
\end{tabular}
\end{table}

\addtolength{\textheight}{-2.1cm}
\section{Discussion and future work}

The results shown in \secref{sec:exp} illustrate that with the proposed pipeline, it was possible to localize an \ac{uav} with respect to a map without requiring a downward-looking camera. We will consider a few key findings and illustrate the most important avenues for potential future research.

Even by using a simple classical image matching criterion, and despite the differences in appearance in map and \ac{uav} images, we were able to localize an \ac{uav} with very modest requirements for initialization to a mean error of approximately less than 50 meters \ac{rms} error after convergence in $(x, y)$ translation after a travel distance of approximately 2 kilometers. In areas where there is less ambiguity in the map (such as above an area with roads), mean localization error is reduced significantly, to a level of less than 20 m (\figref{fig:error_plots}). In areas with ambiguities in map matching, the system is able to provide a measure of ambiguity (standard deviation) and track multiple hypotheses that are consistent with odometry measurements until the \ac{uav} again reaches an area with distinguishable features. In all datasets, localization error is less than that of an odometry-only localization, unless there is significant appearance difference of the area over which the \ac{uav} is flying to the dataset which was used for characterizing a matching score, or the terrain appearance is naturally ambiguous.

This suggests that engineering a suitable matching criteria is a key enabler in image-based localization, to improve robustness against appearance change and to improve its ability to separate between correct and incorrect matches also in areas where less pronounced visual features appear. Also, the speed of convergence of the \ac{mcl} filter and its robustness to errors is tied to its capability to distinguish between correct and incorrect pose candidates. In conclusion, a matching method that provides a small overlapping coefficient over a wide range of different appearance conditions is key. Another key requirement for low localization error is that the matching function should provide a distinct peak, whose width is in proportion to odometry error, in vicinity of the correct pose. Engineering a matching score that meets these requirements is a potential line of future research.

Instead of assuming that the entire environment in which the \ac{uav} is operating is flat, which is common in other works on this topic, we made a less strict assumption that the area observed within a batch is planar. Our assumption is still overly simplistic for scenarios where the \ac{uav} is flying at low altitudes relative to the extent of altitude variations. Potential future research directions include investigating the gap between downward-facing images and orthoprojected camera images to quantify the need for more elaborate reconstruction means, potentially solving structure of the environment from camera motion, and producing orthoprojection using the constructed 3D model.

Even though we demonstrated the operation of our localization solution with three separate datasets and the reference maps were captured several months before the UAV imagery, the imaging conditions were good in all the datasets. It can be expected that a classical area-based matching method would not suffice in the case of more significant illlumination or appearance change.

\section{Conclusions}

We proposed an end-to-end solution to localize an \ac{uav} using visual-inertial odometry and orthorectification of an \ac{uav} image, thus mitigating the need to have a dedicated downward-facing camera. We defined a way to quantify the goodness of match of an \ac{uav} orthoimage to an orthophoto map. We used a map and three experimental datasets of \ac{uav} images acquired with different sensors and at different times to demonstrate that the localization solution for camera pose converges to an acceptable localization error level even when initial pose information is very inaccurate.

Appearance variations across seasons, time-of-day, and weather conditions create a great limitation for matching camera images to maps, which presents a major challenge for absolute visual localization using orthophotos or satellite images. A crucial missing link is a matching measure that is invariant to those variations. Future work is needed to address the development of such measures, that would allow methods such as the one presented here to perform reliably in the wild.
\bibliographystyle{./bibliography/IEEEtran}
\bibliography{./bibliography/IEEEabrv,./bibliography/bibliography}

\begin{thebibliography}{10}
\providecommand{\url}[1]{#1}
\csname url@samestyle\endcsname
\providecommand{\newblock}{\relax}
\providecommand{\bibinfo}[2]{#2}
\providecommand{\BIBentrySTDinterwordspacing}{\spaceskip=0pt\relax}
\providecommand{\BIBentryALTinterwordstretchfactor}{4}
\providecommand{\BIBentryALTinterwordspacing}{\spaceskip=\fontdimen2\font plus
\BIBentryALTinterwordstretchfactor\fontdimen3\font minus
  \fontdimen4\font\relax}
\providecommand{\BIBforeignlanguage}[2]{{%
\expandafter\ifx\csname l@#1\endcsname\relax
\typeout{** WARNING: IEEEtran.bst: No hyphenation pattern has been}%
\typeout{** loaded for the language `#1'. Using the pattern for}%
\typeout{** the default language instead.}%
\else
\language=\csname l@#1\endcsname
\fi
#2}}
\providecommand{\BIBdecl}{\relax}
\BIBdecl

\bibitem{7747236}
C.~{Cadena}, L.~{Carlone}, H.~{Carrillo}, Y.~{Latif}, D.~{Scaramuzza},
  J.~{Neira}, I.~{Reid}, and J.~J. {Leonard}, ``Past, present, and future of
  simultaneous localization and mapping: Toward the robust-perception age,''
  \emph{IEEE Transactions on Robotics}, vol.~32, no.~6, pp. 1309--1332, Dec
  2016.

\bibitem{COUTURIER2021103666}
\BIBentryALTinterwordspacing
A.~Couturier and M.~A. Akhloufi, ``A review on absolute visual localization for
  uav,'' \emph{Robotics and Autonomous Systems}, vol. 135, 2021.
\BIBentrySTDinterwordspacing

\bibitem{9341682}
J.~{Choi} and H.~{Myung}, ``Brm localization: Uav localization in gnss-denied
  environments based on matching of numerical map and uav images,'' in
  \emph{2020 IEEE/RSJ International Conference on Intelligent Robots and
  Systems (IROS)}, 2020, pp. 4537--4544.

\bibitem{10.1117/12.2518516}
A.~Couturier and M.~A. Akhloufi, ``{UAV navigation in GPS-denied environment
  using particle filtered RVL},'' in \emph{Situation Awareness in Degraded
  Environments 2019}, J.~J.~N. Sanders-Reed and J.~T. J.~A. III, Eds., vol.
  11019.\hskip 1em plus 0.5em minus 0.4em\relax SPIE, 2019, pp. 188 -- 198.

\bibitem{MANTELLI2019304}
\BIBentryALTinterwordspacing
M.~Mantelli, D.~Pittol, R.~Neuland, A.~Ribacki, R.~Maffei, V.~Jorge,
  E.~Prestes, and M.~Kolberg, ``A novel measurement model based on abbrief for
  global localization of a uav over satellite images,'' \emph{Robotics and
  Autonomous Systems}, vol. 112, pp. 304--319, 2019.
\BIBentrySTDinterwordspacing

\bibitem{schleiss2019translating}
M.~Schleiss, ``Translating aerial images into street-map-like representations
  for visual self-localization of uavs,'' \emph{Int. Arch. Photogramm. Remote
  Sens. Spat. Inf. Sci}, pp. 575--578, 2019.

\bibitem{7418753}
M.~{Shan}, F.~{Wang}, F.~{Lin}, Z.~{Gao}, Y.~Z. {Tang}, and B.~M. {Chen},
  ``Google map aided visual navigation for uavs in gps-denied environment,'' in
  \emph{2015 IEEE International Conference on Robotics and Biomimetics
  (ROBIO)}, Dec 2015, pp. 114--119.

\bibitem{6942633}
H.~{Chiu}, A.~{Das}, P.~{Miller}, S.~{Samarasekera}, and R.~{Kumar}, ``Precise
  vision-aided aerial navigation,'' in \emph{2014 IEEE/RSJ International
  Conference on Intelligent Robots and Systems}, 2014, pp. 688--695.

\bibitem{6521919}
B.~{Grelsson}, M.~{Felsberg}, and F.~{Isaksson}, ``Efficient 7d aerial pose
  estimation,'' in \emph{IEEE Workshop on Robot Vision (WORV)}, 2013, pp.
  88--95.

\bibitem{6943040}
A.~{Yol}, B.~{Delabarre}, A.~{Dame}, J.~{Dartois}, and E.~{Marchand},
  ``Vision-based absolute localization for unmanned aerial vehicles,'' in
  \emph{2014 IEEE/RSJ International Conference on Intelligent Robots and
  Systems}, Sep. 2014, pp. 3429--3434.

\bibitem{5354165}
K.~{Son}, Y.~{Hwang}, and I.~{Kweon}, ``Uav global pose estimation by matching
  forward-looking aerial images with satellite images,'' in \emph{2009 IEEE/RSJ
  International Conference on Intelligent Robots and Systems}, 2009, pp.
  3880--3887.

\bibitem{9196606}
B.~{Patel}, T.~D. {Barfoot}, and A.~P. {Schoellig}, ``Visual localization with
  google earth images for robust global pose estimation of uavs,'' in
  \emph{2020 IEEE International Conference on Robotics and Automation (ICRA)},
  2020, pp. 6491--6497.

\bibitem{8794228}
A.~{Shetty} and G.~X. {Gao}, ``Uav pose estimation using cross-view
  geolocalization with satellite imagery,'' in \emph{2019 International
  Conference on Robotics and Automation (ICRA)}, May 2019, pp. 1827--1833.

\bibitem{8575361}
A.~{Nassar}, K.~{Amer}, R.~{ElHakim}, and M.~{ElHelw}, ``A deep cnn-based
  framework for enhanced aerial imagery registration with applications to uav
  geolocalization,'' in \emph{2018 IEEE/CVF Conference on Computer Vision and
  Pattern Recognition Workshops (CVPRW)}, June 2018.

\bibitem{8793558}
H.~{Goforth} and S.~{Lucey}, ``Gps-denied uav localization using pre-existing
  satellite imagery,'' in \emph{2019 International Conference on Robotics and
  Automation (ICRA)}, May 2019, pp. 2974--2980.

\bibitem{ThrunProbabilisticRobotics05}
S.~Thrun, W.~Burgard, and D.~Fox, \emph{Probabilistic Robotics (Intelligent
  Robotics and Autonomous Agents)}.\hskip 1em plus 0.5em minus 0.4em\relax The
  MIT Press, 2005.

\bibitem{shi1994good}
J.~Shi \emph{et~al.}, ``Good features to track,'' in \emph{IEEE conference on
  computer vision and pattern recognition (CVPR)}.\hskip 1em plus 0.5em minus
  0.4em\relax IEEE, 1994, pp. 593--600.

\bibitem{bouguet2001pyramidal}
J.-Y. Bouguet \emph{et~al.}, ``Pyramidal implementation of the affine lucas
  kanade feature tracker description of the algorithm,'' \emph{Intel
  corporation}, vol.~5, no. 1-10, p.~4, 2001.

\bibitem{opencv}
G.~Bradski, ``{The OpenCV Library},'' \emph{Dr. Dobb's Journal of Software
  Tools}, 2000.

\bibitem{titterton2004strapdown}
D.~Titterton, J.~Weston, J.~Weston, I.~of~Electrical~Engineers, A.~I.
  of~Aeronautics, and Astronautics, \emph{Strapdown Inertial Navigation
  Technology}.\hskip 1em plus 0.5em minus 0.4em\relax Institution of
  Engineering and Technology, 2004.

\bibitem{woodman2007introduction}
O.~J. Woodman, ``An introduction to inertial navigation,'' University of
  Cambridge, Computer Laboratory, Tech. Rep., 2007.

\bibitem{ADIS16488Adatasheet}
\emph{ADIS16488A Ten Degrees of Freedom Inertial Sensor Datasheet}, Analog
  Devices, 2018, rev. F.

\bibitem{doi:10.1177/0278364906072768}
\BIBentryALTinterwordspacing
F.~Dellaert and M.~Kaess, ``Square root sam: Simultaneous localization and
  mapping via square root information smoothing,'' \emph{The International
  Journal of Robotics Research}, vol.~25, no.~12, pp. 1181--1203, 2006.
\BIBentrySTDinterwordspacing

\bibitem{huber1964}
\BIBentryALTinterwordspacing
P.~J. Huber, ``Robust estimation of a location parameter,'' \emph{Ann. Math.
  Statist.}, vol.~35, no.~1, pp. 73--101, 03 1964.
\BIBentrySTDinterwordspacing

\bibitem{gtsam}
{Contributors of GTSAM project}, ``{GTSAM Library}.''

\bibitem{HARTLEY1997146}
\BIBentryALTinterwordspacing
R.~I. Hartley and P.~Sturm, ``Triangulation,'' \emph{Computer Vision and Image
  Understanding}, vol.~68, no.~2, pp. 146 -- 157, 1997.
\BIBentrySTDinterwordspacing

\bibitem{doi:https://doi.org/10.1002/9780470316849}
D.~W. Scott, \emph{Multivariate Density Estimation}.\hskip 1em plus 0.5em minus
  0.4em\relax John Wiley \& Sons, Ltd, 1992.

\bibitem{Roma2002}
N.~Roma, J.~Santos-Victor, and J.~Tomé, \emph{A Comparative Analysis of
  Cross-Correlation Matching Algorithms Using a Pyramidal Resolution
  Approach}.\hskip 1em plus 0.5em minus 0.4em\relax World Scientific, 2002, pp.
  117--142.

\end{thebibliography}

\end{document}